\documentclass[final]{cvpr}

\usepackage{times}
\usepackage{epsfig}
\usepackage{graphicx}
\usepackage{amsmath}
\usepackage{amssymb}
\usepackage{multirow}
\usepackage{subfig}
\usepackage{xspace}

\newcommand{\x}{\mathbf{x}}
\newcommand{\f}{\mathbf{f}}
\newcommand{\p}{\mathbf{p}}
\newcommand{\muvec}{\boldsymbol{\mu}}
\newcommand{\sigmavec}{\boldsymbol{\sigma}}

\usepackage[pagebackref=true,breaklinks=true,colorlinks,bookmarks=false]{hyperref}

\begin{document}
\newcommand{\reid}{Re-ID\xspace}

%%%%%%%%% TITLE
\title{Learn by Guessing: Multi-Step Pseudo-Label Refinement for\\ Person Re-Identification}

\author{Tiago de C. G. Pereira, Teofilo E. de Campos\\
Universidade de Brasília - UnB\\
Brasília-DF, Brazil\\
{\tt\small pereira.tiago@aluno.unb.br, t.decampos@oxfordalumni.org }
%\and
%Teofilo E. de Campos\\
%Universidade de Brasília - UnB\\
%Brasília-DF, Brazil\\
%{\tt\small t.decampos@oxfordalumni.org}
}

\maketitle

%%%%%%%%% ABSTRACT
\begin{abstract}
    Unsupervised Domain Adaptation (UDA) methods for person Re-Identification (Re-ID) rely on target domain samples 
    to model the marginal distribution of the data. 
    To deal with the lack of target domain labels, 
    UDA methods leverage
    information from labeled source samples and unlabeled 
    target samples. A promising approach relies on the use of unsupervised learning as part of the pipeline, such as clustering methods. 
    The quality of the clusters clearly plays a major role in methods performance, but this point has been overlooked. %  by the majority of methods. 
    In this work, we propose a multi-step pseudo-label refinement method to select the best possible clusters and keep improving them so that these clusters become closer to the class divisions without knowledge of the class labels.
    Our refinement method includes a cluster selection strategy and a camera-based normalization
    method which reduces the within-domain variations caused by the use of multiple cameras in person \reid.
    This allows our method to reach state-of-the-art UDA results on DukeMTMC$\rightarrow$Market1501 (source$\rightarrow$target). 
    We surpass state-of-the-art for UDA \reid by $3.4\%$ on Market1501$\rightarrow$DukeMTMC datasets, which is a more challenging adaptation setup because
    the target domain (DukeMTMC) has eight distinct cameras.
    Furthermore, the camera-based normalization method causes a significant
    reduction in the number of iterations required for training convergence.
\end{abstract}

%%%%%%%%% BODY TEXT
\section{Introduction}

Person re-identification (Re-ID) aims at matching person images from 
different non-overlapping cameras views. This is an essential feature
for diverse real word challenges, such as smart cities \cite{smart_cities}, 
intelligent video surveillance \cite{WANG20133}, suspicious action recognition 
\cite{surveillance} and pedestrian retrieval \cite{SVDNet}.

With all these popular possible applications, there is a clear demand for
robust \reid systems in the industry. 
% the industry necessity for robust \reid systems is well established. 
% Furthermore, 
Academic research groups have achieved 
remarkable in-domain results on popular person \reid datasets such as Market1501 
\cite{zheng2015scalable} and DukeMTMC-reID~\cite{zheng2017unlabeled}. 
Despite these advances, there is still a 
dichotomy between the success in academic results versus the 
industrial application. This is because the best academic results 
\eg \cite{stReid,Luo_2019_CVPR_Workshops,Zhou_2020_CVPR}
are based on supervised methods that require
a huge amount of annotated data for their training. 
Gathering data is not a problem 
nowadays, as CCTV systems are omnipresent. However, annotating 
images is a very expensive and tedious task that requires a lot of manual work. 

% Therefore, a lot of unlabeled data is available and we want to have a robust 
% \reid system. 
% One could think of using a pre-trained state-of-the-art \reid model, 
The use of pre-trained state-of-the-art \reid models usually leads to 
disappointing results because each group of cameras has distinct 
characteristics, such as illumination, resolution, noise level, orientation, pose, distance,
focal length, amount of people's motion as well as factors that influence
the appearance of people, such as ethnicity, type of location (\eg leisure vs work places)
and weather conditions.
% Then, models trained on a source domain tend to perform poorly on a new domain with totally different
% cameras and characteristics.

Some methods have been proposed to reduce this gap and unlock \reid systems for real world problems. 
That is the case of domain invariant models \cite{Style_Norm,Song_2019_CVPR,Frustratingly},
which have the ambitious goal of being applicable to any domain even if no samples are given
from some of the potential target domains. Although domain invariance is indeed
the key to creating a widely applicable method, such methods usually do not outperform
methods in which unlabeled target domain samples are given. Since gathering unlabeled samples
is a virtually effortless task, the need for domain invariant methods is not so urgent.
A common alternative is the use of Generative Adversarial Networks (GANs) to align domains, 
allowing the model to perform well in a target domain even if supervised training is 
only performed in the source domain  \cite{AD_Cluster,Zhong_2018_CVPR,Deng_2018_CVPR,Adaptative_transfer}. 
In addition, there are Unsupervised Domain Adaptation (UDA) methods
which have been achieving notable results in cross-domain person \reid.
These methods typically rely on a process of target domain pseudo-labels generation.
This allows them to use actual target domain images without previous annotation
\cite{Lin_2020_CVPR,Ge2020Mutual,HCT,dg_net,reid_visapp_2020,ProgressiveLearning,SSG}. 

In this work, we dive deep in the UDA \reid setup utilizing pseudo-labels to enhance
models performance in target domain. The quality of pseudo-labels clearly is essential 
for the performance of this kind of method. However, pseudo-labels are expected to be
noisy in this scenario. Many methods have use soft
cost functions to deal with this noise, however we 
believe that cleaning and improving % and asserting better 
pseudo-labels is key to achieve high 
performance. We therefore focus our work in two main points: 
camera-based normalization, which we observed to be key to reduce domain variance;
and a % Nao gostei do uso de "intelligent", por isso removi em todo o artigo, mas ficou super boring.
% O IDEAL SERIA INVENTAR UM NOME OU SIGLA PRA ESSA TECNICA DE CLUSTER SELECTION
% E TALVEZ ATE MENCIONA-LA NO TITULO.
novel clusters selection strategy. The latter removes outlying clusters
and generate pseudo-labels with important characteristics to 
help model convergence. This strategy aims to generate clusters which
are dense and each contain samples of one person captured from the view
of multiple cameras.

Enhancing cluster quality has been overlooked by % the majority of
methods based on pseudo-labels and this has certainly held back many methods.
% based methods and it hold them back.
To evaluate our proposal we work with the most 
popular cross-domain dataset 
in unsupervised \reid works: Market1501 and DukeMTMC. 
Our main contribution is 
a multi-step pseudo-label refinement that keeps cleaning and 
improving the predicted target domain label space to enhance model performance 
without the burden of annotating data. Further to proposing
a new pipeline, we introduce strategies to build and select clusters
in a way that maximizes the model's generalization ability and
its potential to transfer learning to new \reid datasets where the
labels are unknown.
Our method achieves UDA \reid state-of-art for DukeMTMC $\rightarrow$ 
Market1501 and significantly pushes state-of-the-art for Market1501 
$\rightarrow$ DukeMTMC, improving results in $3.4\%$ w.r.t.\ the
best results we are aware of.
We achieve state-of-the-art results without any post-processing methods, however
we are aware that re-ranking algorithms are helpful for metric learning tasks. We thus evaluate our model using k-reciprocal encoding re-ranking \cite{Zhong_2017_CVPR} and improve our results by further $2.1\%$ and $2.9\%$ for DukeMTMC and Market1501, respectively.

\section{Related works}
Person \reid has been a trending computer vision research topic. % lately. 
There are two main directions for person \reid research: 
\textbf{a)} supervised person \reid, that aims at 
creating the best possible models for in-domain \reid and 
\textbf{b)} unsupervised domain adaptation (UDA) \reid 
focusing on the \reid task in which a model trained in a 
source dataset is adapted to another dataset, where the
labels are not known. The latter is sometimes referred to
as cross-domain \reid.
In this field, each person \reid dataset has images captured
from multiple cameras and the dataset as a whole is  assumed to be 
one domain. Although domain adaptation techniques can be applied
within a single dataset, i.e., to adapt samples from one camera
view to another, we focus on the problem of adapting between 
different datasets. This setting is more related
to a real system deploying setting because training can
be done using a dataset containing several viewpoints, but
the deployment scenario is a different set of data
where it is easy to capture unlabeled samples but labeled
samples are not available.

\textbf{Generalizable person \reid} \cite{Style_Norm,Song_2019_CVPR,Frustratingly,zhuang2020rethinking} 
pursue models that are 
domain invariant, so they can perform well in diverse \reid datasets without 
the need of any adaptation. That is an interesting approach for real world \reid 
challenges, although it still does not perform as well as in-domain person \reid.
To achieve a domain invariant model, Zhuang \etal \cite{zhuang2020rethinking}
propose to replace all batch normalization layers of a deep CNN by camera batch normalization
layers. These layers learn to apply batch normalization for each camera reducing the 
within-domain camera variance, this also helps the model to learn camera invariant features 
that are more robust to domain changes. % evaluation. 
Jin \etal \cite{Style_Norm} propose a Style Normalization and Restitution (SNR) 
module that firstly alleviates camera style variations and then 
restores identity relevant features that may have been discarded in the style 
normalization step, reducing the influence of camera style change on feature vectors.

\textbf{GAN-based person \reid} \cite{AD_Cluster,Zhong_2018_CVPR,Deng_2018_CVPR,Adaptative_transfer,person_transfer_GAN,dg_net} 
have been widely used to reduce 
the domain gap in \reid datasets. Deng \etal \cite{Deng_2018_CVPR} use  
cycleGAN to transfer the style from an unlabeled target domain to a labeled source 
domain, so they leverage source domain annotations to apply their trained model
to images that are more similar to the source % target 
domain ones. 
Zhai \etal \cite{AD_Cluster} used GANs to augment 
the target domain training data, so they could create images that preserved the 
person ID and that simulates other camera % dataset % CHECK IF IT REALLY MAKES SENSE TO REPLACE CAMERA BY DATASET HERE!
views at the same time. 

\textbf{Pseudo-Labels Generation for Person \reid} \cite{reid_visapp_2020,Ge2020Mutual,HCT,SSG,dg_net,AD_Cluster,ProgressiveLearning,Lin_2020_CVPR}
predict the label space for an unlabeled target domain, assume those predictions
are correct and use then to fine-tune a model previously trained on source domain.
This approach has shown remarkable results and is the idea behind % responsible 
current state-of-the-art UDA \reid methods. 
The drawback with pseudo-labels is that if the domains are not similar 
enough, they can lead to negative transfer, because the labeling
noise might be too high. To deal with that, Ge \etal \cite{Ge2020Mutual} propose a 
soft softmax-triplet loss to leverage from pseudo-labels without overfitting 
their model. Zeng \etal \cite{HCT} propose a hierarchical clustering method 
to reduce the influence of outliers and use a batch hard triplet loss to 
% approximate the
bring outliers closer to interesting regions so they could be used later on. 
% O QUE SAO "INTERESTING REGIONS"???

We believe that a generalize-able \reid model is pre-requisite for a strong UDA 
method. For this reason we adopt IBN-Net50-a as our backbone and apply a camera guided 
feature normalization in target domain to reduce the domain gap. We also understand 
the importance of cleaning the pseudo-labels, which is what motivate us to a clustering algorithm
with outlier detection and to propose a clustering selection step to feed our model only 
with data that is predicted to be more reliable. 
Next section details our methods.
% Our methods will be further discussed in section  Section \ref{methodology}.

%% Ainda não consegui preparar uma imagem do metodo para colocar aqui 
% VAI SER BOM COLOCAR MESMO, O ARTIGO NAO ESTA NADA VISUAL...

\section{Methodology}
\begin{figure*}[htpb]%
    \centering
    \includegraphics[width=.9\textwidth]{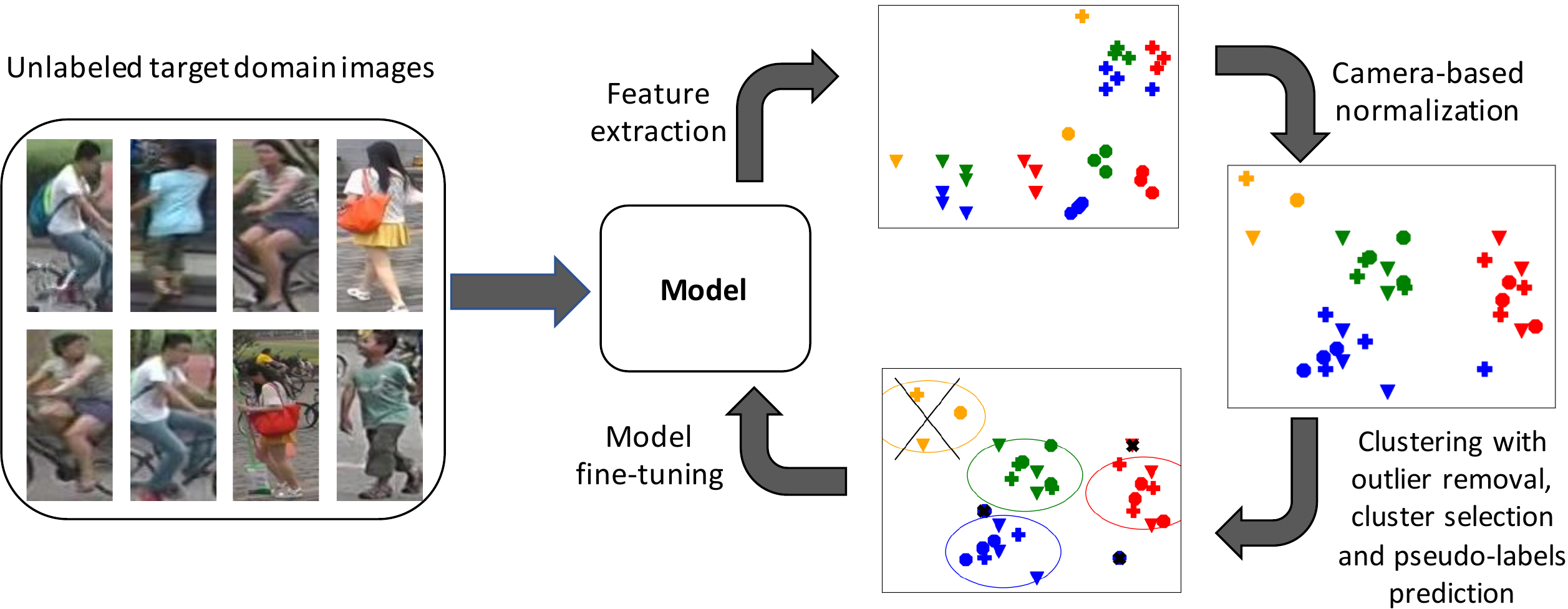}
    \caption{Our Multi-Step Pseudo-Label Refinement pipeline. The proposed method consists of four components: extraction of features from unlabeled target domain images, camera-based normalization, prediction of pseudo-labels with a density-based clustering algorithm, selection of reliable clusters and fine-tuning of the model. The pipeline is cyclical, because at each step it predicts more robust pseudo-labels that offer new information for the model. In the feature space panels, each shape (\eg triangle, plus signal and circle) represent a camera view and each color represent a person ID.}%
    \label{fig:pipeline}%
\end{figure*}

\label{methodology}
In this Section we present and discuss all steps that compose our method in deep details.
% , then we break this Section and 6 Subsections. In Subsection 
In \S \ref{backbone} we discuss the commonly used backbones for \reid
and in % Subsection 
\S \ref{PL} we present 
the concept of progressive learning. Then, we % will see 
review some clustering techniques
in \S \ref{Clustering} and how to generate % confident 
robust clusters in \S \ref{Cherry}
and \ref{CN}. 
Finally, in \S \ref{Our} we explain how to effectively combine all techniques 
in our training protocol. 

\subsection{Backbone}
\label{backbone}
When working in cross domain tasks, the model generalization % capacity
ability
is key to success.
% to achieve great results. Also, 
Normalization techniques have a very important role for that. % to improve model generalization.

Nowadays, the typical \reid system relies on ResNet \cite{resnet50} as their backbone
(usually the ResNet-50 model),%  and this
which is a safe choice, because \reid is a task that requires multiple semantic levels 
to produce robust embeddings and the residual blocks help to propagate these multiple 
semantic levels to deeper layers. 
Also, ResNet is a well studied CNN that lead to a step change
in the performance on the ImageNet dataset \cite{imagenet_cvpr09}.

However, the vanilla ResNet has its generalization compromised because it does not
include instance-batch normalization.
% lacks normalization when compared to other models like IBN-Net50-a
In \cite{IBN}, the authors proposed  IBN-Net50, which replaces batch normalization (BN) layers with instance batch
normalization (IBN) layers. The IBN-Net carefully integrates IN and BN as building blocks,
significantly increasing its generalization ability.
% and has shown better generalization capacity compared to ResNet. Therefore, 
For this reason, we choose it as our backbone. More specifically, we use IBN-Net50-a,
which offers a good compromise between model size and performance.

\subsection{Progressive Learning}
\label{PL}
Progressive learning is an iterative technique proposed by Hehe \etal \cite{ProgressiveLearning}
% which consist in a cycle that repeats until model convergence. The cycle is 
composed of three parts: 
\textbf{a)} generating target domain pseudo-labels to train the model without
labeled data, 
\textbf{b)} fine-tuning the model with the previous generated pseudo-labels and 
\textbf{c)} evaluating the model. 
This set of steps is iterated until convergence.
This approach relates to some classical UDN and Transductive Transfer Learning techniques, such as 
\cite{farajidavar_etal_ATTM_bookchapter2017} and \cite{Long_2013_ICCV} for standard classification tasks.

To get full advantage of progressive learning it is important to generate new pseudo-labels at each iteration, so the model will have new stimulus to keep learning. Therefore it is important that % need to guarantee that 
in each step the new pseudo-labels get closer of the real labels. 
However, if the initial model is not good enough, this leads to negative transfer \cite{PAN_YANG_TL}
and the performance of the system actually degrades as it iterates.
However, since target labels are unknown, it is not possible to predict
negative transfer. 
% So, it is important to use only reliable data as pseudo labels, that is data that your model is confident. Then, at each iteration the model will be performing better and will be able to select more data as pseudo-labels.

For this reason, we argue that progressive learning must be coupled
with other techniques, such as the method we describe in the next 
sections, particularly in \S \ref{Cherry} and \ref{CN}.
In those sections, we propose to 
evaluate the reliability of samples and their pseudo-labels
based on the confidence of the model. If only reliable samples and their
pseudo-labels are used, the model should progressively improve and
generate more robust pseudo-labels in the consecutive iterations.

\subsection{Clustering techniques}
\label{Clustering}
For standard classification tasks, pseudo-labels generation is direct: it is
assumed that the predictions obtained are usually correct and these predictions
on the target set are used as pseudo class labels. 
However, due to the lack of control on the number of classes, person \reid is usually 
approached as a metric learning task. The model prediction is therefore not a label,
but a feature vector in a space where samples of the same person are expected to
lie closer to each other (and further to samples of different people). 
Therefore, it is necessary to use clustering algorithms 
and define each cluster as a pseudo-label (or pseudo person ID).

Therefore, given a target domain $\mathcal{D}^t$ with $N$ images $\{\x_i \}_{i=1}^N$
we need to predict their labels $\{ y_i \}_{i=1}^N$. Then, we use a model pre-trained 
on source domain $\mathcal{D}^s$ to extract the features for each image $\{\x_i \}_{i=1}^N$ from 
$\mathcal{D}^t$ and use a clustering algorithm to group/predict each image label\footnote{We 
use bold math symbols for vectors.}.

\subsubsection{K-means}
\label{kmeans}
As a first choice, we used the k-means \cite{kmeans} algorithm to cluster our data. The 
only parameter k-means need is the number of clusters $k$. For our experiments, we choose 
$k$ using this heuristic:
\begin{equation}
    k = \left \lfloor \frac{N}{15} \right \rfloor ~ ,
    \label{eq:k_mean}
\end{equation}
where $N$ is the total number of training images in target domain $\mathcal{D}^t$.
If all clusters have a balanced number of features (images) this would mean that we are 
assuming that each person ID in the target domain contains about $15$ samples.

There are two problems with k-means for \reid: \textbf{a)} how to define $k$ without information 
about $\mathcal{D}^t$ and \textbf{b)} as stated by Zeng \etal\cite{HCT} k-means does not 
have an outlier detector, so the outliers may drag the centroids away from 
denser % interesting 
regions, causing the decision boundaries to shift, potentially cutting through sets of 
samples that actually belong to the same people.

\subsubsection{DBSCAN}
\label{dbscan}
As discussed above, k-means is not recommended to generate robust pseudo-labels for UDA \reid
methods. Therefore, we propose the usage of DBSCAN \cite{DBSCAN} which is a density-based 
clustering algorithm designed to deal with large and noisy databases.

In a Domain Adaptation \reid scenario we can say that the hard samples are actually noise, % represent the noise, 
so a clustering algorithm that % detects these samples and predicts 
itentify
then as outliers is fundamental
to improve results. Furthermore, when applying Progressive Learning we can leave 
hard samples out for some iterations and bring then to the pseudo-labeled dataset in later % advanced
iterations where our model is stronger  %better 
and the level of confidence in those hard samples is higher.

% This outlier detection method therefore deals with the problem of k-means, 
% problem of the outliers dragging the clusters
% away from interesting regions, 
% but what about the number of clusters? 

One important point is that DBSCAN does not require a pre-defined number of clusters
(as in k-means), but it requires two parameters: the maximum distance between 
two samples to determine them as neighbors 
($\epsilon$) and the minimum number of samples to consider a region as dense 
($\textit{min\_{}s}$).

% PLEASE CHECK THIS PARAGRAPH:
In our experiments, we set $\textit{min\_{}s}=4$. 
As for the parameter $\epsilon$, its value depends on the spread 
of the data. We performed a simple search in an early training step
determine a value that would balance the number of clusters selected 
and the number of outliers. This lead to $\epsilon=0.35$ when DukeMTMC 
is the target domain and $\epsilon=0.42$ when Market1501 is the target domain.

\subsection{Cluster Selection} 
\label{Cherry}
\reid datasets have disjoint label spaces, that is given a source domain $\mathcal{D}^s$
and a target domain $\mathcal{D}^t$ their labels space do not share the same classes, \ie 
\begin{equation}
    \{ y_i \}^s \neq \{ y_j \}^t \ \forall \ \textit{i,j} .
\end{equation}
As mentioned before, the usual way to deal with this is by approaching 
\reid as a metric-learning task. 

Therefore, \reid methods typically use triplet loss with batch hard \cite{defense} and 
PK sampling. The PK sampling method consist in selecting P identities with K samples from 
each identity to form a mini-batch in training stage, which leads to the following:
\begin{equation}
    \textit{batch\_size} = P \times K .
\end{equation}

In this work we used the triplet loss and PK sampling to train our models, so we expect that 
every person ID has at least K images. This clustering step therefore ignores clusters 
with less than K images.

An important factor for \reid models is to learn features that are robust to camera 
view variations. For that we guarantee that, %propose to use guarantee that, 
in the training stage,
our model is fed with samples of the same person ID in different cameras. 
Therefore, we also prune clusters that had images from only one camera view.

\subsection{Camera-Guided Feature Normalization}
\label{CN}
The high variance present in \reid datasets is mainly caused by the different camera views,
as each view has its own characteristics. This is why a model trained in a source dataset
presents poor results when evaluated in a target dataset (or domain). 
Normally, \reid models learn robust 
features for known views, but lack the ability to generalize for new unseen views. 

Pereira and de~Campos \cite{reid_visapp_2020} realize that this lack of generalization 
power has a negative impact in pseudo-labels generation. They point that the
main reason for that is the fact that, in new unseen cameras, the model tends to cluster images 
by cameras rather than clustering images from the same 
person in different views. The majority of clusters would therefore
be ignored in the Cluster Selection step. 

Zhuang \etal \cite{zhuang2020rethinking} replaced all batch normalization layers by 
camera batch normalization layers. Although this helped them to reduce the data variance 
between camera views, they normalize the data only on the source domain.
We propose to run this camera feature normalization step before the pseudo-labels step on 
the target domain training set. By generating pseudo-labels that are 
normalized by camera information, our method guides the model to learn robust features in the 
target domain space without the need of changing the model architecture or having additional
cost functions. 

Camera-guided feature normalization therefore aims to reduce the target domain variance, 
enhance the model capacity in the target domain and create better pseudo-labels that 
further will result in a more robust model.

To apply camera guided feature normalization, we first divide all target domain 
training images $\{ \x_i \}^t$ in $c$ groups where $c$ is the number of cameras views in
the dataset. Then we extract their features $\f_{i}^{(c)}$ with our model and 
calculate, for each camera $c$, its mean $\muvec^{(c)}$ and its standard deviation
$\sigmavec^{(c)}$. Finally, each feature is normalized by 
\begin{equation}
    \overline{\f}_{i}^{(c)} = \frac{\f_{i}^{(c)} - \muvec^{(c)}}{\sigmavec^{(c)}} .
    \label{eq:norm}
\end{equation}
The normalized features $\overline{\f}_{i}^{(c)}$ are then used to generate the pseudo-labels.

\subsection{Our Training protocol}
\label{Our}

\subsubsection{Baseline}
\label{Baseline}
First of all, we train our model in the source domain $\mathcal{D}^s$ as a baseline. All 
our models use the IBN-Net50-a as backbone and outputs feature vectors $\f$ with 2048 dimensions and
a logit prediction vectors $\p$.

Our loss function has three components: \textbf{a)} a batch hard triplet loss 
($\mathcal{L}_{\textit{tri}}$) \cite{defense} that maps $\f$ in an Euclidean vector space, 
\textbf{b)} a center loss ($\mathcal{L}_{c}$) \cite{center_loss}
to guarantee cluster compactness and \textbf{c)} and a cross entropy label smooth
loss ($\mathcal{L}_{\textit{ID}}$) \cite{label_smooth} that uses the logit vectors $\p$ to 
learn a person ID classifier. The smoothed person ID component has been proved to help \reid systems
\cite{Luo_2019_CVPR_Workshops} even though the training IDs are disjoint from the testing IDs.
Furthermore, its soft labels has shown interesting features for UDA \reid 
\cite{Ge2020Mutual}. Our loss function is thus given by Equation \ref{eq:loss}.
\begin{equation}
    \mathcal{L} = \mathcal{L}_{\textit{tri}} + \mathcal{L}_{\textit{ID}} + 0.005 \mathcal{L}_{c}
    \label{eq:loss}
\end{equation}
% HOW DID YOU COME UP WITH THE WEIGHT 0.005. IS THIS USED IN OTHER PAPERS?
The weight given to the cross entropy loss is the same that was used in \cite{Luo_2019_CVPR_Workshops}.

We start our training with pre-trained weights from ImageNet \cite{DeCAF} and 
use the Adam optimizer for $90$ epochs with a warm-up learning rate scheduler
defined by Equation \ref{eq:lr}, which is based on \cite{Luo_2019_CVPR_Workshops}. 
\begin{equation}
  lr =
    \begin{cases}
        3.5\times10^{-5} \times \frac{\textit{epoch}}{10} &, ~ \textit{epoch} \leq 10 \\ 
        3.5\times10^{-4} &, ~ 10 < \textit{epoch} \leq 40\\ 
        3.5\times10^{-5} &, ~ 40 < \textit{epoch} \leq 70 \\ 
        3.5\times10^{-6} &, ~ \textit{epoch} > 70
    \end{cases}       
    \label{eq:lr}
\end{equation}
For data augmentation we use random erase \cite{randomCrop},
resize images to $256\times128$ and apply a random color transformation that could be a $20\%$ 
brightness variation or a $15\%$ contrast variation. We also use random horizontal flipping.

\subsubsection{Usupervised Domain Adaptation}
For unsupervised domain adaptation, we start with the model pre-trained  
in $\mathcal{D}^s$ and use it to extract all the features $f$ from $\mathcal{D}^t$ training images. Once we have all these features extracted, we separate them by 
camera and use Equation~\ref{eq:norm} to normalize them. Then, we use DBSCAN 
to create general clusters in $\mathcal{D}^t$ and finally apply our  cluster selection strategy to keep only the clusters which are potentially the
the most reliable ones.

From the selected clusters we create our pseudo-labeled dataset and 
use it to fine-tune our previous model. 
Since the domains are different datasets, the person IDs on the pseudo-labeled 
dataset are always different from those of the source dataset.
Additionally, as our progressive learning strategy iterates, pseudo-labels are expected to change.
Therefore, it is expected that the cross-entropy loss 
$\mathcal{L}_{\textit{ID}}$ spikes in first iterations, which
can destabilize the training process and lead to catastrophic 
forgetting. To prevent that, we follow the transfer learning
strategy of freezing the body of our 
model for 20 epochs and let the last fully connected layer 
learn a good enough $\p$.
Then, we unfreeze our model and complete the fine-tuning 
following the procedure described in \ref{Baseline}.

% NAO ENTENDI ESSE TRECHO ACIMA:
% "learn a good enough $\p$ to control the $\mathcal{L}_{\textit{ID}}$." 
% A IDEIA EH MANTER A CROSS ENTROPY SUB CONTROLE?
% EU ENTENDI QUE ESSE FINE TUNING SEPARA CORPO DE CABECA (E ESCREVI DE ACORDO),
% (O QUE EH UMA TECNICA PADRAO EM FINE TUNING). ENTENDI CORRETAMENTE?

After the fine-tuning we evaluate our model on $\mathcal{D}^t$ % and if the Mean Average Precision increased from before the fine-tuning, 
and iterate the whole process, according to the progressive learning strategy.
% ISSO TEM ETICA DUVIDOSA, MAS POR ENQUANTO PREFIRO NAO DEIXAR TAO EXPLICITO QUE VOCE USA ROTULOS DO DESTINO PARA FAZER ADAPTACAO. UMA COISA QUE PODERIA SER FEITA EH, INVES DE USAR mAP COMO CRITERIO DE PARADA, SERIA SIMPLESMENTE DEIXAR O METODO CONVERGIR, I.E. SE A QUANTIDADE DE PSEUDO-LABELS PARAR DE MUDAR DE UMA ITERACAO PARA A OUTRA, VOCE PARA.   
\section{Experiments}
\subsection{Datasets}
We performed our experiments on the Market1501 and DukeMTMC datasets, interchanging then as 
source and target domains to analyze our domain adaptation method. Following the
standard in this field, we used 
cumulative matches curve (CMC) and mean average precision (mAP) as evaluation 
metrics.

\textbf{Market1501} is an outdoors dataset containing images across 6 cameras views where 
each person appears in at least 2 different cameras. In total there are 
32668 images being 12936 images from 751 identities for training and 19732 images 
from 750 identities for testing.

\textbf{DukeMTMC} was also built using outdoor cameras. It contains 36411
images from 1404 identities. Those images were split as 16522 for training, 2228 for 
query and 17661 for test gallery.

\begin{table*}[htpb]
    \centering
    \begin{small}
    \begin{tabular}{|c|cccc|cccc|}
    \hline
    \multirow{2}{*}{\textbf{Methods}} & \multicolumn{4}{c|}{\textbf{Market1501 $\rightarrow$ DukeMTMC}} & \multicolumn{4}{c|}{\textbf{DukeMTMC $\rightarrow$ Market1501}} \\ \cline{2-9}
                                      &  Rank-1 &  Rank-5 &  Rank-10 & mAP  &  Rank-1 &  Rank-5 &  Rank-10 & mAP  \\ \hline
    Supervised               &  82.4   &  92.0   &  94.5    & 68.8 &  91.2   &  92.0   &  98.4    & 79.2 \\ \
    Direct Transfer                   &  44.7   &  60.7   &  66.4    & 27.3 &  58.9   &  74.3   &  80.1    & 29.0 \\ \    
    \textbf{Ours}                     &  82.7   &  90.5   &  93.5    & 69.3 &  89.1   &  95.8   &  97.2    & 73.6 \\ \hline
    \end{tabular}
    \end{small}
    \caption{Comparison of our results with results using supervised learning on the target domain 
    (which is expected to give the best results) and 
    direct transfer results, i.e.\ the use of a model trained on source directly applied to the target
    domain, without domain adaptation (which is expected to be a lower bound).}
    \label{table:ours}
\end{table*}

\begin{table*}[htpb]
    \centering
    \begin{small}
    \begin{tabular}{|c|cccc|cccc|}
    \hline
    \multirow{2}{*}{\textbf{Methods}}       & \multicolumn{4}{c|}{\textbf{Market1501 $\rightarrow$ DukeMTMC}} & \multicolumn{4}{c|}{\textbf{DukeMTMC $\rightarrow$ Market1501}} \\ \cline{2-9}
                                            &  Rank-1 &  Rank-5 &  Rank-10 & mAP  &  Rank-1 &  Rank-5 &  Rank-10 & mAP  \\ \hline
    SPGAN \cite{Deng_2018_CVPR}             &  46.9   &  62.6   &  68.5    & 26.4 &  58.1   &  76.0   &  82.7    & 26.9 \\ \
    UCDA-CCE \cite{UCDA}                    &  55.4   &    -    &    -     & 36.7 &  64.3   &    -    &    -     & 34.5 \\ \    
    ARN \cite{ARN}                          &  60.2   &  73.9   &  79.5    & 33.4 &  70.3   &  80.4   &  86.3    & 39.4 \\ \
    MAR \cite{MAR}                          &  67.1   &  79.8   &   -      & 48.0 &  67.7   &  81.9   &     -    & 40.0 \\ \
    ECN \cite{ECN}                          &  63.3   &  75.8   &  80.4    & 40.4 &  75.1   &  87.6   &  91.6    & 43.0 \\ \
    PDA-Net \cite{PDA-Net}                  &  63.2   &  77.0   &  82.5    & 45.1 &  75.2   &  86.3   &  90.2    & 47.6 \\ \
    EANet \cite{EANet}                      &  67.7   &   -     &    -     & 48.0 &  78.0   &    -    &    -     & 51.6 \\ \
    CBN \cite{zhuang2020rethinking} + ECN   &  68.0   &  80.0   &  83.9    & 44.9 &  81.7   &  91.9   &  94.7    & 52.0 \\ \
    Theory \cite{Theory}                    &  68.4   &  80.1   &  83.5    & 49.0 &  75.8   &  89.5   &  93.2    & 53.7 \\ \
    CR-GAN \cite{CR-GAN}                    &  68.9   &  80.2   &  84.7    & 48.6 &  77.7   &  89.7   &  92.7    & 54.0 \\ \
    PCB-PAST \cite{PCB-PAST}                &  72.4   &    -    &    -     & 54.3 &  78.4   &    -    &    -     & 54.6 \\ \
    AD Cluster \cite{AD_Cluster}            &  72.6   &  82.5   &  85.5    & 54.1 &  86.7   &  94.4   &  96.5    & 68.3 \\ \
    SSG \cite{SSG}                          &  76.0   &  85.8   &  89.3    & 60.3 &  86.2   &  94.6   &  96.5    & 68.7 \\ \
    DG-Net++ \cite{dg_net}                  &  78.9   &  87.8   &  90.4    & 63.8 &  82.1   &  90.2   &  92.7    & 61.7 \\ \
    MMT \cite{Ge2020Mutual}                 &  \textit{79.3}   &  \textit{89.1}   &  \textit{92.4}    & \textit{65.7} &  \underline{90.9}   &  \textbf{96.4}   &  \textbf{97.9}    & \underline{76.5} \\ \hline
    \textbf{Ours}                           &  \underline{82.7}   &  \underline{90.5}   &  \textbf{93.5}    & \underline{69.3} &  \textit{89.1}   &  \underline{95.8}   &  \underline{97.2}    & \textit{73.6} \\ \
    \textbf{Ours + Re-Ranking \cite{Zhong_2017_CVPR}}                           &  \textbf{84.8}   &  \textbf{90.8}   &  \underline{93.2}    & \textbf{81.2} &  \textbf{92.0}   &  \textit{95.3}   &  \textit{96.6}    & \textbf{88.1} \\ \hline
    \end{tabular}
    \end{small}
    \caption{Comparison of our results with state-of-art methods in UDA. We highlighted in bold, underline and italic the first, second and third best results, respectively.}
    \label{table:SOTA}
\end{table*}

%\subsection{Generalizable Efficiency}

%% Repensar essa seção, talvez retreinar os nossos modelos de direct transfer sem 
%% o uso de random crop, pois sabemos que isso danifica o poder de generalização

%\begin{table*}[t]
%    \centering
%    \begin{tabular}{|c|cccc|cccc|}
%    \hline
%    \multirow{2}{*}{\textbf{Methods}} & \multicolumn{4}{c|}{\textbf{Market1501 $\rightarrow$ DukeMTMC}} & \multicolumn{4}{c|}{\textbf{DukeMTMC $\rightarrow$ Market1501}} \\ \cline{2-9}
%                                      &  Rank-1 &  Rank-5 &  Rank-10 & mAP  &  Rank-1 &  Rank-5 &  Rank-10 & mAP  \\ \hline
%    CBN \cite{zhuang2020rethinking}   &  58.7   &  74.1   &  78.1    & 38.2 &  72.7   &  85.8   &  90.7    & 43.0 \\ \
%    SNR \cite{Style_Norm}             &  55.1   &  -      &    -     & 33.3 &  66.7   &    -    &    -     & 33.9 \\ \hline
%    \textbf{Ours}                     &  52.6   &  69.2   &  75.0    & 33.1 &  60.7   &  77.3   &  83.3    & 30.1 \\ \hline
%    \end{tabular}

%    \caption{Comparasion of our direct transfer + camera guided feature normalization 
%    against state-of-art methods in domain generalizable task.}
%\end{table*}

\subsection{Comparison with Supervised Learning and Direct Transfer}
In Table \ref{table:ours} we compare our baseline results to the direct transfer and 
to our proposed method. 
The supervised learning was done using samples and labels from the 
target domain training samples. Since samples and labels are from
the same domain as the test set, this is expected to give results 
that are better than those of domain adaptation settings.
The aim of the supervised learning experiments is to
understand the capacity of the model in each dataset.
% NEXT SENTENCE IS ACTUALLY TALKING ABOUT DIRECT TRANSFER!
% Then, we used the baseline model trained in DukeMTMC and 
% evaluated it in Market1501 (direct transfer), and vice-versa.

The Direct transfer method is used to evaluate the domain shift and the model 
generalization power. It is expected that this setting gives
resutls that are worse than the domain adaptation setting, because no
knowledge of the target set is used in the training process.
Our method does not focus on being generalizable, we aim to 
use the source domain knowledge as start and enhance the model's performance 
in target domain without any labels. We found it important to
present direct transfer results in order to show how much our 
method enhances over it.

As one can see, our method reaches remarkable results for 
DukeMTMC as a target dataset.
It can be surprising to see that we even 
surpass the supervised result in $0.3\%$ and $0.5\%$ for CMC rank-1 and 
mAP, respectively. DukeMTMC is a dataset with a high intra-variance caused by its 
eight distinct camera views. We believe that the camera-guided normalization
applied before the clustering step provided pseudo-labels that were more robust 
to camera view variations. Therefore, the method was able to learn camera 
invariant features. 
It is also likely that by transferring from one dataset to another,
our method was less prone to over-fitting than the supervised learning
setting. 

For Market1501 as a target, our method performed equally well enhancing the direct transfer 
result in $30.2\%$ and $44.6\%$ for CMC rank-1 and mAP, respectively. However, with 
lower intra-variance in Market1501 the supervised result is already 
saturated. Therefore, even tough labels from the target set were not used,
our methods gives results which are not far below those
of the supervised setting.

\subsection{Comparison with state-of-art UDA results}

In Table \ref{table:SOTA} we compare our multi-step pseudo-label refinement method 
with multiple state-of-the-art \reid UDA methods. As one can see, we beat all other methods 
in DukeMTMC target dataset and push the state-of-the-art by $3.4\%$ and $3.6\%$ for CMC rank-1 and mAP,
respectively. For Market1501 we are able to reach second place with a noticeable 
gap to the third place with an improvement of $2.4\%$ and $4.9\%$ for CMC rank-1 and mAP,
respectively.

In addition, our framework have a lightweight architecture when compared to other frameworks that 
achieve state-of-the-art. 
MMT \cite{Ge2020Mutual} uses two CNNs so that one generates soft labels for the other. DG-Net++ \cite{dg_net} uses a extremely complex framework with GANs and multiple encoders and decoders. 

%% Acho que deveria escrever mais aqui, mas estou sem ideias

As we approach \reid as a metric learning task, re-ranking algorithms have a great impact in the results. Then, we evaluated our model using k-reciprocal encoding re-ranking \cite{Zhong_2017_CVPR} which combines the original distance with the Jaccard distance in an unsupervised manner. The importance to use a ranking system is shown with an CMC Rank-1 improvement of $2.1\%$ for DukeMTMC and $2.9\%$ in Market1501 when compared to our raw method. Also, re-ranking significantly pushes the mAP performance in $11.9\%$ for DukeMTMC and $14.5\%$ for Market1501.

\subsection{Ablation studies}
% \subsubsection{Techniques contributions}

\begin{table}[htpb]
    \centering
    \begin{small}
    \begin{tabular}{|l|cc|cc|}
    \hline
    \multirow{2}{*}{\textbf{Methods}} & \multicolumn{2}{c|}{\textbf{M $\rightarrow$ D}} & \multicolumn{2}{c|}{\textbf{D $\rightarrow$ M}} \\ \cline{2-5}
                        &  Rank-1 & mAP  &  Rank-1 & mAP  \\ \hline
    ResNet 50 \cite{Luo_2019_CVPR_Workshops}  &   41.4     &   25.7  &    54.3    &   25.5  \\ \
    + IBN-Net50-a       &  44.7   & 27.3 &  58.9   & 29.0 \\ \
    + DA                &  52.2   & 37.1 &  60.1   & 34.8 \\ \
    + PL                &  52.2   & 37.1 &  61.4   & 35.5 \\ \
    + Cluster Selection &  77.2   & 61.8 &  86.5   & 66.0 \\ \
    + CN                &  82.7   & 69.3 &  89.1   & 73.6 \\ \hline
    \end{tabular}
    \end{small}
    \caption{The contribution of each method in the model performance evaluated on 
    Market1501 and DukeMTMC-reID datasets. PL means Progressive Learning, CN 
    stands for Camera Guided Normalization and DA for Domain Adaptation}
    \label{table:ablation}
\end{table}

Table \ref{table:ablation} shows how each technique contributes to our final 
method performance. 

\textbf{IBN-Net50-a: } the difference between the original Resnet-50 and the IBN-Net50-a
is that the IBN-Net50-a modifies all batch normalization layers so they also take 
advantage of an instance normalization. This modification enhances the model 
normalization capacity and generalization power, leading to an improvement on the CMC rank-1 
performance improvement of $3.3\%$ in DukeMTMC and $4.6\%$ in Market1501.

\textbf{Domain Adaptation: } in Table \ref{table:ablation} we call as domain adaptation 
the use of pseudo-labels from target domain for training. This clustering guided 
domain adaptation allows our model to train using actual images from target domain, 
which facilitates the model to learn various aspects of the domain, such as illumination,
camera angles, person pose. Learning the characteristics from target domain 
is a major factor for domain adaptation which becomes evident by the
CMC rank-1 improvement of $7.5\%$ in DukeMTMC and $1.2\%$ in Market1501.

\textbf{Progressive Learning: } this technique has a great potential to keep improving
the model's performance with new pseudo-labels. %  organization. 
However, as we said in Section \ref{PL} to get full advantage of this technique 
ones need to guarantee that the pseudo-labels are close to the class divisions.
Therefore, this step only gives a significant improvement if associated
with the proposed cluster selection technique. 
% and if we do not 
% perform an intelligent selection of the clusters we can not guarantee it.
In Table \ref{table:ablation} results, the progressive learning results were 
obtained using the raw clusters defined by the clustering algorithm. Then, the model used 
all the available information in target domain and overfitted to these pseudo-labels.
In the next step these clusters tend to be the same and the model does not have a 
stimulus to learn better features. This is why the progressive learning results 
on their own do not seem to help.

\textbf{Cluster Selection: } % as discussed for the progressive learning, this type of
this method relies on a continuous improvement on the pseudo-labels. % To guarantee it
for that, we 
remove clusters that are unlikely to help improve the model, such as small 
clusters with less than 4 images and clusters that had images from only one camera 
view. Using this strategy % cluster selection 
we can get full advantage of progressive learning 
and push the model to learn camera view invariant features, since all our pseudo-labels
have samples from multiple camera views. 

The real contribution of the progressive learning technique is shown alongside the 
contribution of the cluster selection strategy, because they are complementary 
techniques. This is certainly the most relevant element of our pipeline,
as it leads to a step change in our performance, enhancing
the rank-1 CMC performance in $25.0\%$ for DukeMTMC and $25.1\%$ for Market1501.

\textbf{Camera Guided Normalization: } learning camera invariant features is essential
for person \reid, because the person appearance may vary for different cameras and 
the model has to deal with all types of variations. 
% (ja falou isso antes:)
% During the training 
%  phase it is important to feed the model with examples from the same person in 
% distinct camera views. 
Since target labels are unknown, when the model extracts features from the target domain, instead of grouping
images by the person that appears in them, the feature vectors tend to cluster
camera viewpoints. 
% This is because 
% the model ends up encoding some camera information in the features space. 
The 
camera guided normalization helps to reduce this camera shift and align the 
features from different cameras. This camera alignment allows the cluster method 
to create better clusters with samples from different cameras. Our cluster 
selection method thus selects more clusters to be part of the pseudo-label dataset.
With this richer and camera invariant pseudo-label dataset, our model has
better samples to learn from and its mAP is improved by 
$7.5\%$ for DukeMTMC and $7.6\%$ for Market1501.

\begin{figure}[htpb]%
    \centering
    \includegraphics[width=.39\textwidth]{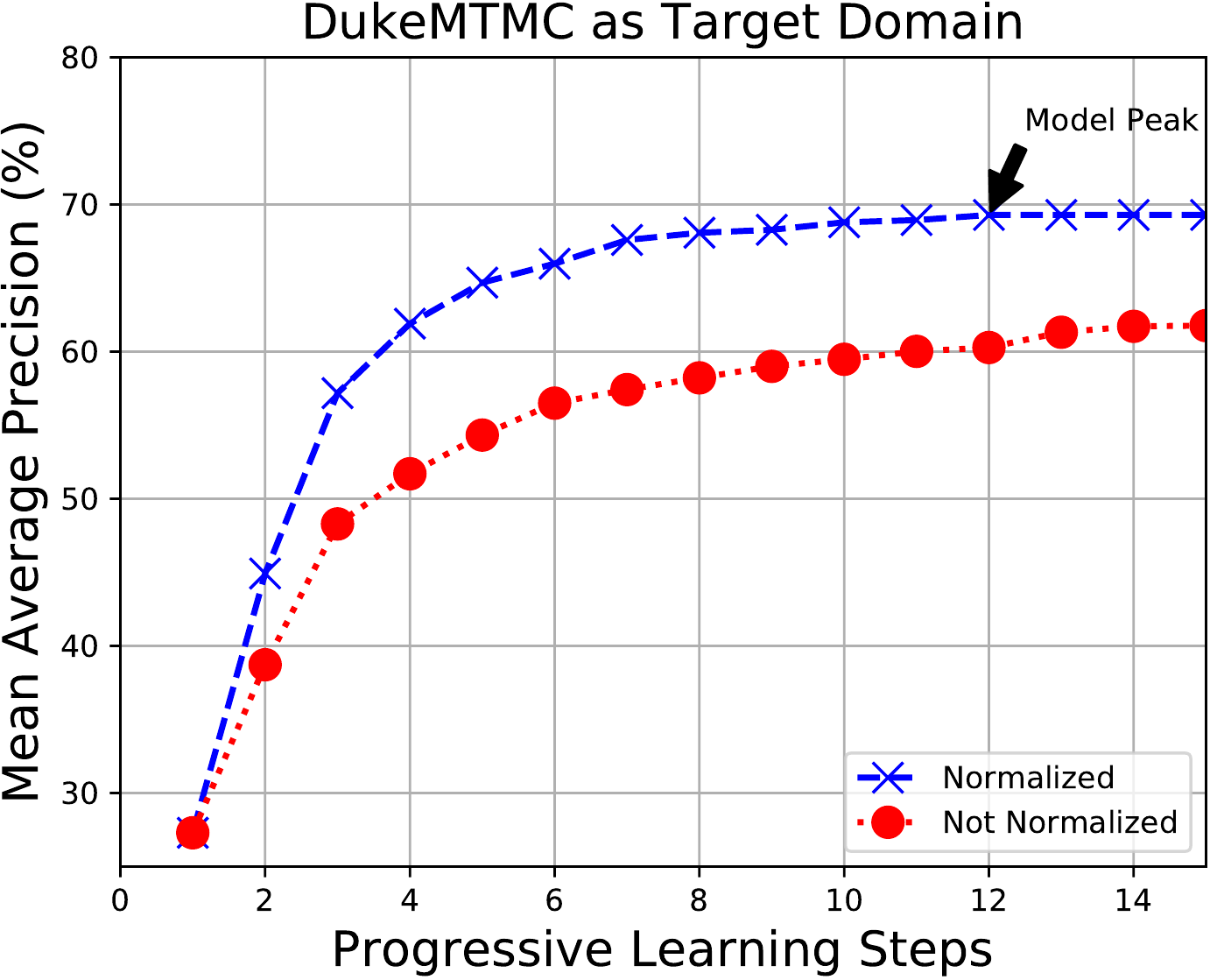}\\
    ~\\
    \includegraphics[width=.39\textwidth]{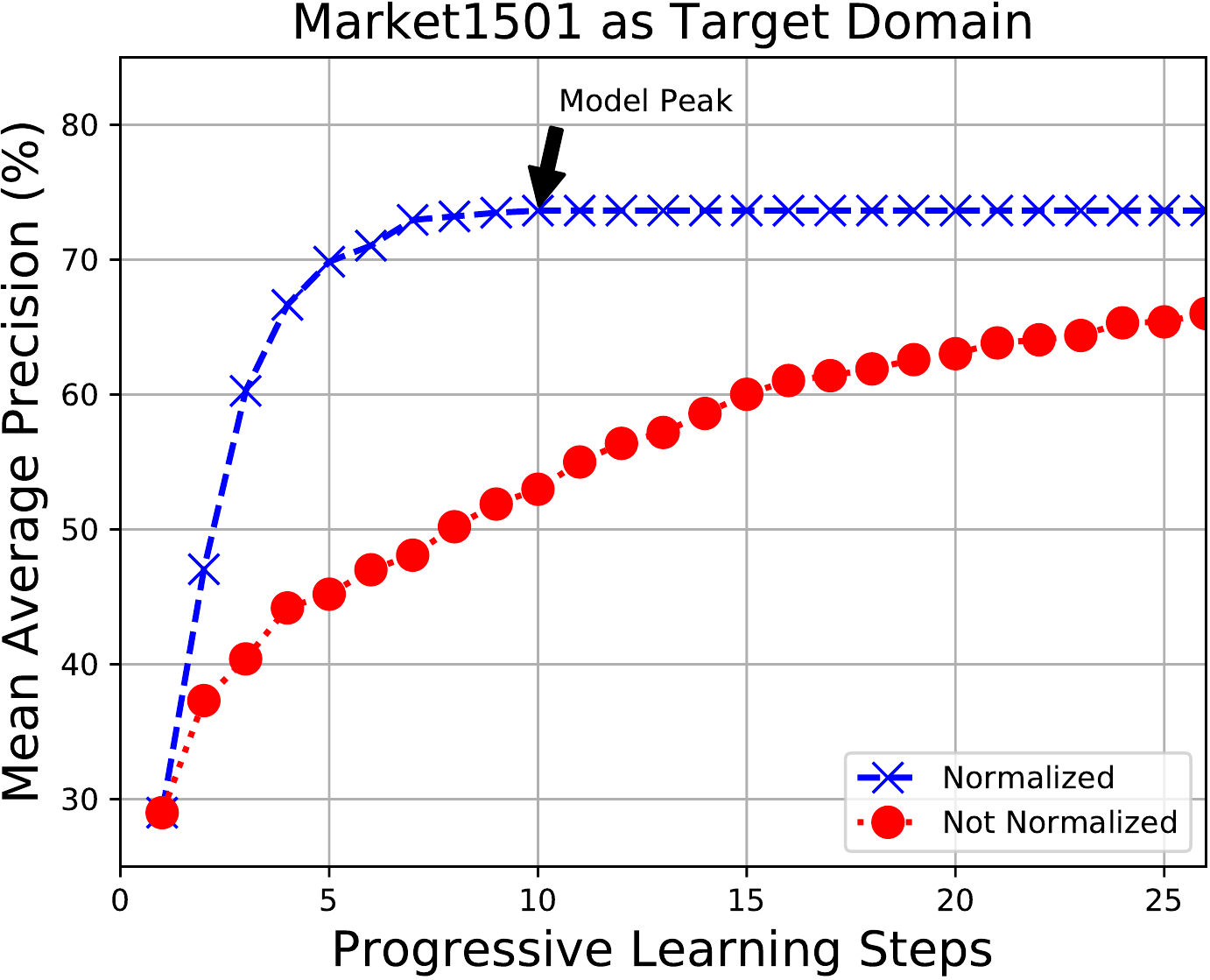}%
    \caption{Comparison of progressive learning steps that take for model convergence
    when using or not camera guided normalization. The black arrow indicates when the 
    model using normalization reached convergence, the points after the black arrow 
    repeat the best model result. The ``not normalized'' curves reach their peak
    at the right end of these plots.}%
    \label{fig:graph}%
\end{figure}

\textbf{Training efficiency:} the better pseudo-labels which are obtained 
when applying camera guided
normalization speeds up the model convergence. Figure \ref{fig:graph}
shows how many progressive learning steps were needed to reach convergence 
with or without camera guided normalization.

%\begin{figure*}[htpb]%
%    \centering
%    
%    \subfloat{\includegraphics[width=0.47\textwidth]{figs/ProgressiveLearning_Duke.pdf}}%
%    \qquad
%    \subfloat{\includegraphics[width=0.47\textwidth]{figs/ProgressiveLearning_Market.pdf}}%
%    
%    \caption{Comparison of progressive learning steps that take for model convergence
%    when using or not camera guided normalization. The black arrow indicates when the 
%    model using normalization reached convergence, the points after the black arrow 
%    repeat the best model result.}%
    
%    \label{fig:graph}%
%\end{figure*}

%\begin{table*}[htpb]
%    \centering
%    \begin{tabular}{|c|ccccc|ccccc|}
%    \hline
%    \multirow{2}{*}{\textbf{Clustering  Alg.}} & \multicolumn{5}{c|}{\textbf{M $\rightarrow$ D}} & \multicolumn{5}{c|}{\textbf{D $\rightarrow$ M}} \\ \cline{2-11}
%                &  Rank-1 &  Rank-5 &  Rank-10 & mAP  & Portion &  Rank-1 &  Rank-5 &  Rank-10 & mAP & Portion  \\ \hline
%    K-Means     &  77.7   & 87.5 & 90.8 & 63.1 & 85.5 &  87.0 & 94.7 &  96.9  & 65.9 & 95.9\\ \
%    DBSCAN      &  82.7   & 90.5 & 93.5 & 69.3 & 69.7 &  89.1 & 95.8 & 97.2  & 73.6 & 79.9\\ \hline
%    \end{tabular}
%
%    \caption{Comparison between DBSCAN and k-means as the clustering algorithm. After cluster selection, different amounts of samples were removed for each clustering method. This portion (in \%) is shown in the last columns.}
%    \label{table:clusters}
%\end{table*}

\begin{table}[htpb]
    \centering
    \begin{small}
    \begin{tabular}{|c|ccccc|}
    \hline
    \multirow{2}{*}{\textbf{Method}} & \multicolumn{5}{c|}{\textbf{M $\rightarrow$ D}} \\ 
    \cline{2-6}
                &  Rank-1 &  Rank-5 &  Rank-10 & mAP  & Portion \\ \hline
    k-means     &  77.7   &    87.5 &     90.8 & 63.1 & 85.5 \\ 
    DBSCAN      &  82.7   &    90.5 &     93.5 & 69.3 & 69.7 \\ \hline
    \multicolumn{6}{c}{~}\\ \hline
    \textbf{Method} & \multicolumn{5}{c|}{\textbf{D $\rightarrow$ M}} \\ 
    \hline
    k-means  &  87.0 & 94.7 &  96.9  & 65.9 & 95.9 \\
    \hline
    DBSCAN&  89.1 & 95.8 & 97.2  & 73.6 & 79.9\\ \hline
    \end{tabular}
    \end{small}
    \caption{Comparison between DBSCAN and k-means as the clustering algorithm. After cluster selection, different amounts of samples were removed for each clustering method. This portion (in \%) is shown in the last columns.}
    \label{table:clusters}
\end{table}

\textbf{Clustering methods:} we ran our multi-step pseudo-label refinement method with two different clustering algorithms in its pipeline: k-means and DBSCAN. Table \ref{table:clusters} presents the results achieved using each of them and also the portion of training data that was selected for use as pseudo-labels after the cluster selection phases.
DBSCAN does not need a fixed number of clusters and has an built-in outlier detector, so it can deal with hard samples better than k-means. For k-means, all samples count, then the hard samples have a negative impact in the quality of the pseudo-labels.
The results in Table~\ref{table:clusters} confirms our hypothesis that it is better to use fewer and less noisy samples.

%Nao consgui fazer uma imagem interessante para mostrar as figuras,pois essa ficou com os numeros muito pequenos. Talvez valha a pena explorar isso nos materiais suplementares onde seria possivel incluir imagens muito grandes sem problema de espaço

%\begin{figure*}[htpb]%
%    \centering
%    
%    \subfloat{\includegraphics[width=0.47\textwidth]{figs/ProgressiveLearning_Duke_cluster.pdf}}%
%    \qquad
%    \subfloat{\includegraphics[width=0.47\textwidth]{figs/ProgressiveLearning_Market_cluster.pdf}}%
%   \caption{Comparison of progressive learning steps that take for model convergence
%    with different clustering algorithms. The percentage numbers written in the graph represent 
%    the portion from training images that were used to preform the iteration. This graph is better
%    visualized in computer.}%
%    
%    \label{fig:cluster}%
%\end{figure*}
\section{Conclusions}
In this work we propose a multi-step pseudo-label refinement method to improve results on Unsupervised Domain Adaptation for Person Re-Identification.
We focus on tackling the problem of having noisy pseudo-labels in this task and proposed a pipeline that reduces the shift caused by camera changes as well as techniques for outlier removal and cluster selection.
Our method includes 
DBSCAN clustering algorithm that was designed to perform well in large and noisy databases; 
a camera-guided normalization step to align features from multiple camera views and allow them to be part of the same clusters;
and a smart cluster selection method that creates optimal pseudo-labels for our training setup and keep improving the pseudo-labels at each progressive learning step.

Our method generates a strong label space for target domain without any supervision. 
We reach state-of-the-art performance on Market1501 as a target dataset and push the state-of-the-art on the challenging DukeMTMC target dataset by $5.5\%$ (or 3.4\% without re-ranking).
Our work hightlights the importance of pseudo-labels refinement with strong normalization techniques. It also takes advantage of a metric learning process and re-ranking \cite{Zhou_2020_CVPR,Zhong_2017_CVPR}. % methods have shown potential to enhance metric learning. 
This combination has clearly proven successul.
% The potential of re-ranking for pseudo-label refinement remain uncovered and we will investigate in the future research.

One possibility for future work is to investigate the use of re-ranking as part
of the clustering step.

\section*{Acknowledgements}
We thank the company that has been sponsoring 
Mr.\ Pereira during most of his MSc studies (further details cannot
be disclosed yet). We are also grateful for the
partial support of Coordenação de Aperfeiçoamento de Pessoal 
de Nível Superior - Brasil (CAPES) - Finance Code 001. 
Dr.\ de Campos received support from the National Council for
Scientific and Technological Development - CNPq - through 
grant PQ~314154/2018-3.
We have also received some support from FAPDF.

{\small
\bibliographystyle{ieee_fullname}
\bibliography{bibliografia}
}

\end{document}